\pdfoutput=1

\documentclass[11pt]{article}

\usepackage[]{EMNLP2022}

\usepackage{times}
\usepackage{latexsym}
\usepackage{amssymb}

\usepackage[T1]{fontenc}

\usepackage[utf8]{inputenc}
\usepackage{subfig}
\usepackage{caption}
\usepackage[english]{babel}
\usepackage{amsthm}

\usepackage{microtype}

\usepackage{inconsolata}

\usepackage{array}

\usepackage{placeins}

\usepackage{tikz-dependency}
\usepackage{algorithm}
\usepackage{algpseudocode}
\usepackage{multirow}
\usepackage{booktabs}

\usepackage{color}
\usepackage{helvet}
\usepackage{textcomp}
\usepackage{booktabs}

\usepackage{color}
\usepackage{graphicx}
\usepackage{amsmath}

\usepackage{hyperref}
\usepackage{url}

\newcommand{\lmtt}[1]{\fontfamily{lmtt}\selectfont{#1}}
\newcommand{\lmss}[1]{\fontfamily{lmss}\selectfont{#1}}

\newcommand{\dat}[1]{\lmss{#1}}

\newcommand{\fsglue}{\dat{FS-GLUE}}
\newcommand{\fsnli}{\dat{FS-NLI}}

\newcommand{\msa}{\metricn_{\text{sensitivity}}}
\newcommand{\mrda}{\metricn_{\text{RDA}}}
\newcommand{\mspread}{\metricn_{\text{\spread{}}}}

\newcommand{\spread}{\lmtt{Spread}}
\newcommand{\figref}[1]{Fig.~\ref{#1}}
\newcommand{\dataset}[1]{\mathcal{D}_{#1}}
\newcommand{\metricn}{h}

\newcommand{\smetric}[2]{h_{#2}(#1)}

\newcommand{\model}{f}
\newcommand{\modeln}[1]{f(#1)}
\newcommand{\train}{\mathcal{D}_{tr}}
\newcommand{\test}{\mathcal{D}_{ts}}

\newcommand{\egg}{e.g.,}
\newcommand{\iee}{i.e.,}
\newcommand{\mfh}{\ensuremath{\mathrm{MFH}}}
\newcommand{\ifh}{\ensuremath{\mathrm{IFH}}}

\newif\ifcomments
\commentstrue
\ifcomments
\usepackage[normalem]{ulem}
\definecolor{CMpurple}{rgb}{0.6,0.18,0.64}
\newcommand\cm[1]{\textcolor{CMpurple}{\textsf{\scriptsize[\textbf{CM\@:} #1]}}}
\newcommand\cmi[1]{\textcolor{CMpurple}{#1}}
\newcommand\cmm[1]{\marginpar{\raggedright\tiny\textcolor{CMpurple}{\textsf{{\bfseries CM\@:} #1}}}}
\newcommand\cms{\bgroup\markoverwith{\textcolor{CMpurple}{\rule[.4ex]{2pt}{0.8pt}}}\ULon}
\else
\newcommand\cm[1]{}
\newcommand\cmi[1]{\ignorespaces}
\newcommand\cmm[1]{}
\newcommand\cms[1]{#1}
\fi

\title{On Measuring the Intrinsic Few-Shot Hardness of Datasets}

\author{First Author \\
  Affiliation / Address line 1 \\
  Affiliation / Address line 2 \\
  Affiliation / Address line 3 \\
  \texttt{email@domain} \\\And
  Second Author \\
  Affiliation / Address line 1 \\
  Affiliation / Address line 2 \\
  Affiliation / Address line 3 \\
  \texttt{email@domain} \\}
  
 \author{
 \textbf{Xinran Zhao\textsuperscript{$\star$}} \quad
 \textbf{Shikhar Murty\textsuperscript{$\star$}}\quad
 \textbf{Christopher D. Manning}\\
 Computer Science Department, Stanford University\\
\texttt{\{xzhaoar,smurty,manning\}@cs.stanford.edu}
}

\begin{document}

\theoremstyle{definition}
\newtheorem{definition}{Definition} 
\maketitle
\renewcommand\thefootnote{}\footnote{\textsuperscript{$\star$} Equal Contribution}

\renewcommand*{\thefootnote}{\arabic{footnote}}
\setcounter{footnote}{0}

\begin{abstract}
While advances in pre-training have led to dramatic improvements in few-shot learning of NLP tasks, there is limited understanding of what drives successful few-shot adaptation in datasets. In particular, given a new dataset and a pre-trained model, what properties of the dataset make it \emph{few-shot learnable} and are these properties independent of the specific adaptation techniques used? We consider an extensive set of recent few-shot learning methods, and show that their performance across a large number of datasets is highly correlated,  showing that few-shot hardness may be intrinsic to datasets, for a given pre-trained model. To estimate intrinsic few-shot hardness, we then propose a simple and lightweight metric called {\spread{}} that captures the intuition that few-shot learning is made possible by exploiting feature-space invariances between training and test samples. Our metric better accounts for few-shot hardness compared to existing notions of hardness, and is \textasciitilde{}8--100x faster to compute.





\end{abstract}

\section{Introduction}

\begin{figure}[t]
    \centering
    \includegraphics[width=\linewidth]{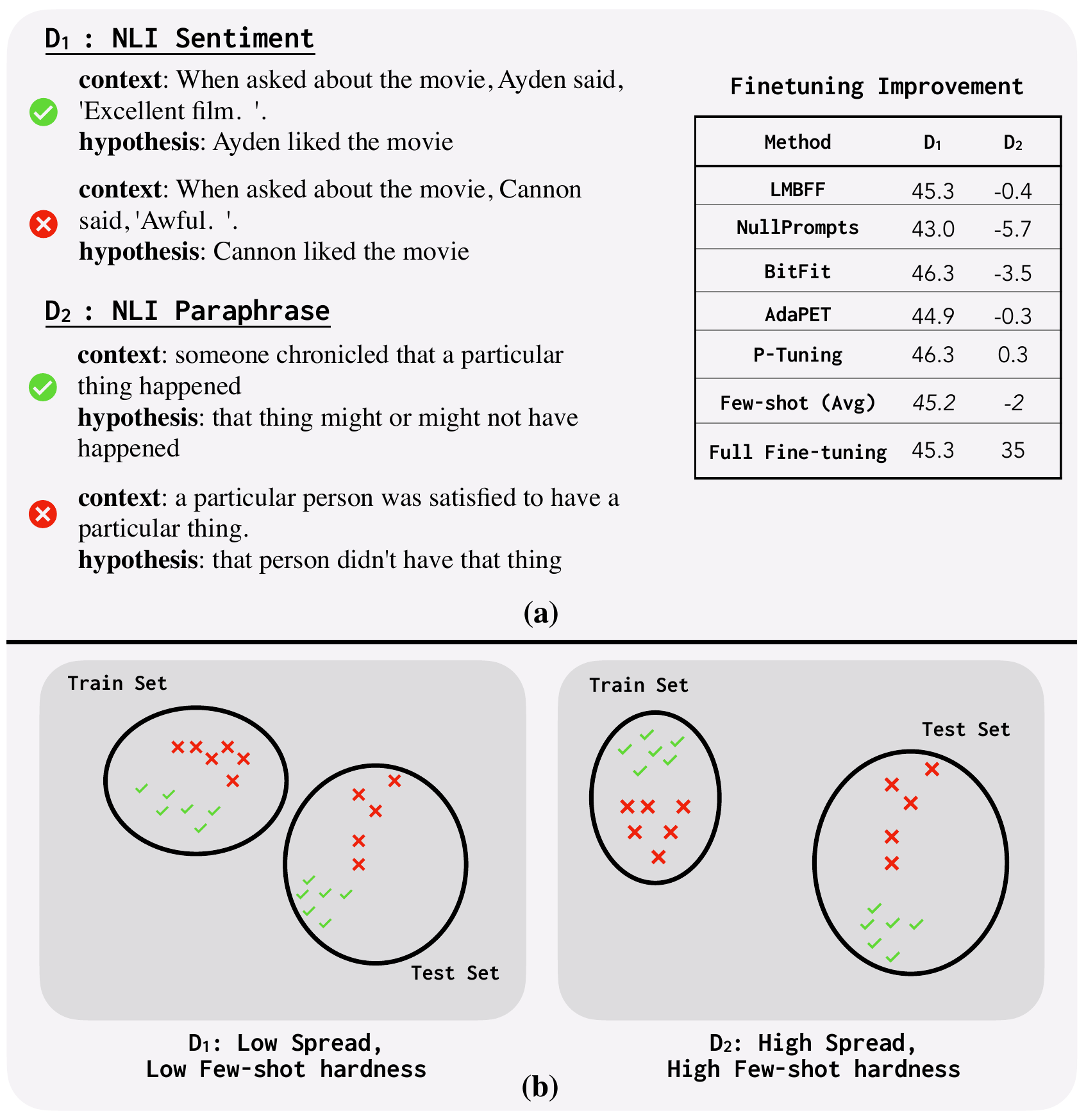}
    \caption{(a) Fine-tuning a model on $D_1$ ($D_2$) on the entire dataset leads to a +45.3 (+35) point improvement over a random baseline. However, most methods are successful at few-shot adaptation on $D_1$ with an average improvement of \textasciitilde{}45 points over a random baseline, while few-shot adaptation is unsuccessful on $D_2$.
    (b) We observe that features of test inputs are closer to training set inputs for $D_1$ than $D_2$, motivating {\spread{}} as a metric for evaluating \emph{few-shot hardness}.}
    \label{fig:spread_intuition}
\end{figure}

A growing body of recent work has shown impressive advances in few-shot adaptation of pre-trained transformers \citep[among others]{radford2019language,schick2020small,DBLP:journals/corr/abs-2005-14165,karimi2021parameterefficient,Liu2021PretrainPA}. Despite this progress, there is no concrete understanding of when and why few-shot learning may be successful for a given pre-trained model. Indeed, \emph{no free lunch} style arguments necessitate the existence of tasks that are not few-shot learnable by a given pre-trained model, regardless of the adaptation method, and in practice, very similar datasets exhibit varying levels of success when state-of-the-art few-shot adaptation methods are applied (\figref{fig:spread_intuition}a). 

This work advances understanding of \emph{few-shot learnability} in two ways. First, we find that, given a dataset, few-shot performance of various adaptation methods is highly correlated. This suggests the existence of adaptation-method independent factors behind few-shot learnability of a dataset, for a given pre-trained model. Next, we propose a simple and lightweight metric to estimate this \emph{intrinsic few-shot learnability}. Concretely, we consider an extensive set of recently proposed adaptation methods and find that on a wide range of datasets, these methods have highly correlated behaviors i.e., the degree to which a few-shot adaptation method succeeds on a dataset is correlated across methods. Next, we consider two recently proposed methods that may be used for assessing \emph{ intrinsic dataset hardness}---Rissanen Data Analysis (RDA, \citet{perez2021rissanen}) and Sensitivity Analysis (SA, \citet{10.1162/tacl_a_00403}). RDA computes dataset hardness as the area under the curve of the test loss as a function of number of training samples, while SA computes hardness by examining how perturbations in the input features cause a model to change its predicted label. From experiments, we show that SA is poorly correlated with few-shot hardness and RDA, while well correlated, is very expensive to compute. In response, we propose a new metric that measures the ability of a model to exploit feature-space invariances between the train and test set to make few-shot generalizations. For instance, consider the datasets in \figref{fig:spread_intuition}b, where $D_1$ has a test set that “looks similar” to the training set while $D_2$ has a test set that looks dissimilar to the training set. We capture this intuition into a lightweight metric called {\spread{}} and show that {\spread{}} correlates as well or better than existing methods (Spearman correlation of 0.467 vs 0.356) while being \textasciitilde{}8--100x more computationally efficient.

\section{Background}

Consider a labeled dataset $\mathcal{D} = \{z^{(1)}, z^{(2)}, \ldots \}$, split into a training set $\train$ and a test set $\test$, where each example $z^{(k)}$ is a tuple consisting of an input $x^{(k)}$ and a label $y^{(k)}$. Typically, in $k$-way $l$-shot learning ($l$ is typically less than 128), $\train$ consists of $l$ examples of each of the $k$ labels. Given some pre-trained model $f$, a few-shot adaptation method $m$ uses $\train$ to modify $f$, outputting an ``adapted model'' that can make predictions on $\test$. 

State-of-the-art approaches for such adaptation typically involve either recasting the task into the pre-training objective of the model \citep{gao2021making, tam2021improving}, or using lightweight / parameter efficient finetuning \citep{houlsby2019parameter,logan2021cutting,li-liang-2021-prefix,DBLP:journals/corr/abs-2110-07602, zaken2021bitfit}. For this work, we experiment with an extensive set of recently proposed few-shot adaptation methods that we further categorize into \emph{prompt-based} methods which includes LMBFF \citep{gao2021making}, AdaPET \citep{tam2021improving}, Null Prompts \citep{logan2021cutting} and Prompt-Bitfit \citep{zaken2021bitfit}\footnote{As demonstrated by \cite{logan2021cutting}, we use trigger prompts in BitFit to improve few-shot performance.}, and \emph{Light-weight} finetuning methods which includes Prefix Tuning \cite{li-liang-2021-prefix} and Compacter \cite{karimi2021parameterefficient}. 

\section{Intrinsic Few-Shot Hardness}
\label{sec:intr_fsh}
\begin{figure}

\centering
\subfloat[]{\label{fig:glue_intrinsic}
\centering
\includegraphics[width=0.47\linewidth]{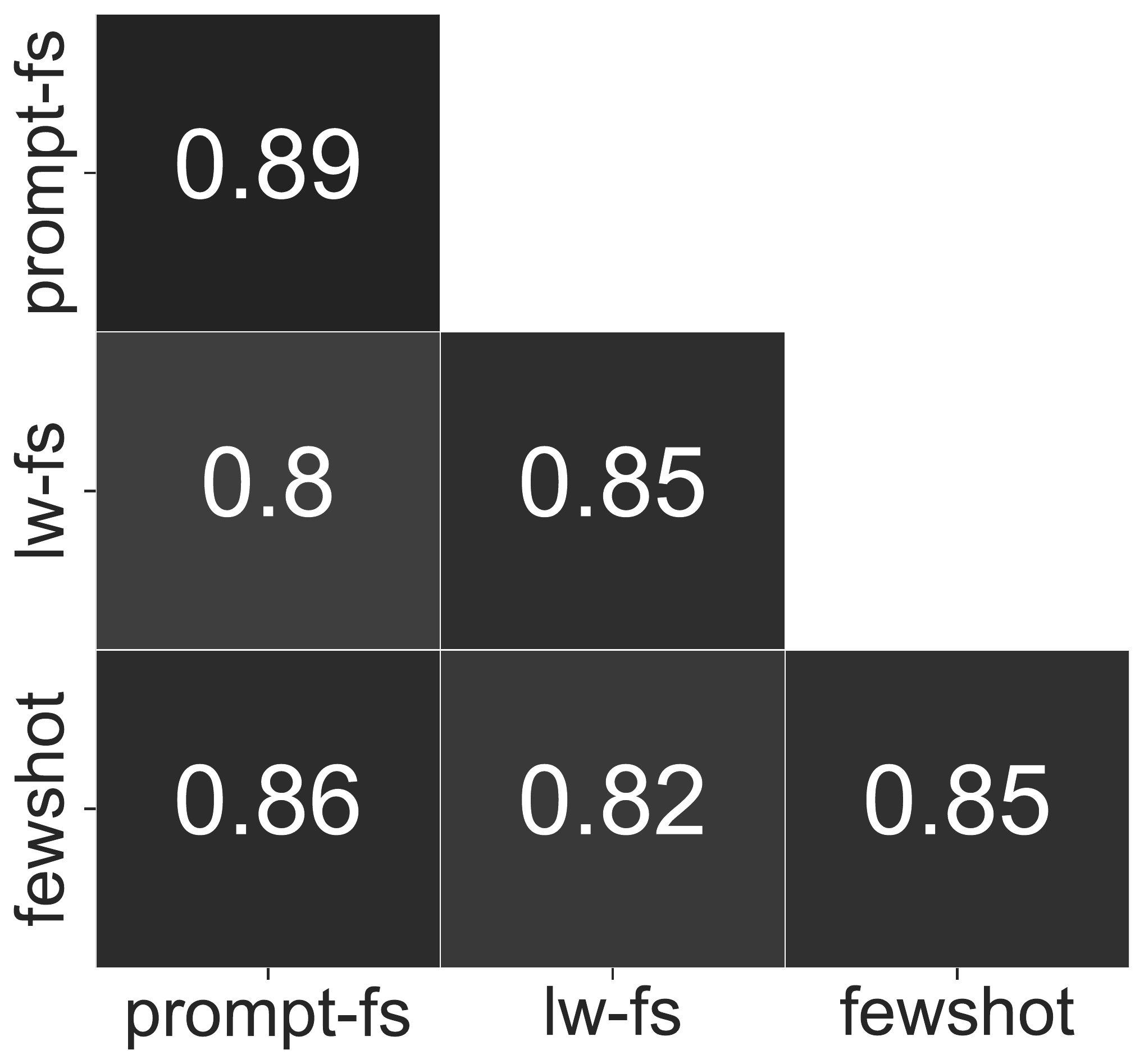}
}
\hfill
\subfloat[]{\label{fig:nli_intrinsic}
\centering

\includegraphics[width=0.47\linewidth]{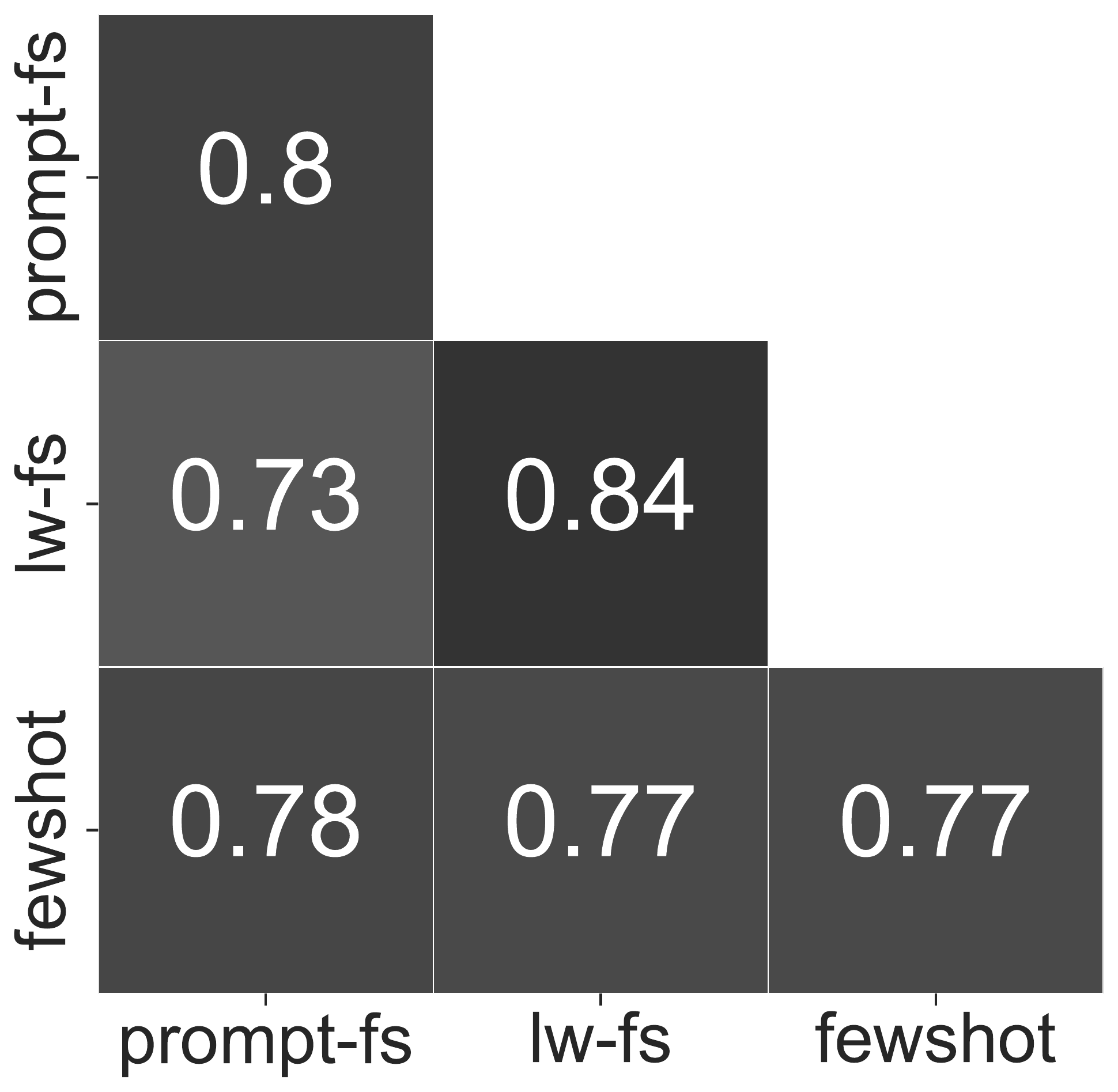}
}
\caption{Average correlation between hardness measurements from various few-shot adaptation methods on (a) {\fsglue{}} and (b) {\fsnli{}}. ``prompt-fs'' refers to prompt-based methods while ``lw-fs'' refers to lightweight finetuning methods. We note a large correlation of 0.85 and 0.77. We also observe a higher correlation among methods from the same category.}
\label{fig:intrinsic_heatmap}
\end{figure}

For a fixed method and dataset, we define ``method specific few-shot hardness'' ($\mfh$) as a function that takes a dataset $\mathcal{D}$ and some adaptation method $m$, and outputs a real number that captures few-shot hardness of $\mathcal{D}$ with respect to $f$ and $m$. Concretely, for this work, we define $\mfh(\mathcal{D}, f, m)$ as the classification accuracy of the adapted model, normalized against the classification accuracy of the majority baseline, on $\test$.

\paragraph{Datasets.} In {\fsglue{}}, we consider 11 tasks (details in Appendix~\ref{sec:dataset_details}) from the GLUE \cite{wang-etal-2018-glue} and SuperGLUE \cite{NEURIPS2019_4496bf24} benchmarks, covering a wide range of task formats and objectives. Additionally, since these tasks have differing textual formats as well as performance metrics (F1 / accuracy / Pearson's correlation), we curate {\fsnli{}} where every dataset is recast into 2 way NLI, and the performance metric is standardized as classification accuracy. To do this, we take datasets from \citet{white-etal-2017-inference,poliak-etal-2018-collecting, DBLP:journals/corr/abs-1909-07521}, giving us a collection of 28 datasets (more details in Appendix~\ref{sec:dataset_details}), covering a wide range of tasks such as sentiment analysis, negation comprehension, name entity classification (NEC), paraphrase detection, event classification, logical entailment etc.

\subsection{Experiments}

We compute $\mfh$ values for all adaptation methods on {\fsglue{}} and {\fsnli{}}. For each dataset, we sample a fixed training set consisting of 64 examples per label. Finally, we control for hyperparameter tuning across methods (details in Appendix~\ref{sec:hyperparam_details}).  We report the average spearman correlation between $\mfh$ values for all pairs of adaptation methods for every dataset in \figref{fig:intrinsic_heatmap}.

\paragraph{Results.} We observe an extremely high average correlation between hardness measurements for different methods---0.85 and 0.77 for {\fsglue{}} and {\fsnli{}} respectively. The average correlation is higher among methods within the same category than among methods from different categories \egg{} on {\fsglue{}}, the average correlation between prompt-based and lightweight finetuning methods is 0.8, while the average correlation among all prompt-based methods is 0.89. We conclude that few-shot hardness may be intrinsic to datasets, \iee{} for a given pre-trained model, a dataset may be ``easy'' or ``hard'' \emph{regardless} of the adaptation method used.

\section{Automatic Metrics For Intrinsic Few-Shot Hardness}
\label{section:measuring_hardness}

Noting that the experiments of Section~\ref{sec:intr_fsh} suggest that few-shot hardness may be intrinsic to datasets, we define the ``intrinsic few-shot hardness'' ($\ifh$) of a dataset as the mean $\mfh$ values across a collection of adaptation method $\mathcal{M} = \{m_1, m_2, \ldots \}$,

\begin{align}
    \ifh(f, \mathcal{D}) \triangleq \frac{\sum_{m \in \mathcal{M}} \mfh(\mathcal{D}, f, m)}{\lvert \mathcal{M} \rvert} 
\end{align}

Clearly, computing $\ifh$ values for some dataset is computationally intensive since it requires running multiple few-shot adaptation methods on $f$. Thus, to automatically estimate $\ifh$, an automatic hardness metric $\metricn$ takes $f$ and $\mathcal{D}$ and outputs a (real-valued) score $\metricn{}(f, \mathcal{D})$ that is well-correlated with $\ifh(f, \mathcal{D})$ across some collection of datasets $\{\dataset{1}, \dataset{2}, \ldots \}$.

\subsection{Existing Metrics} 
We consider two recently proposed dataset hardness metrics that have been applied in the context of pre-trained models. $\msa{}$ ~\cite{10.1162/tacl_a_00403} measures how many tokens of an input need to be perturbed for model predictions to change, on average. A low sensitivity implies that model predictions change only when a large subset of an input's tokens are perturbed, making the dataset ``easy''. $\mrda{}$ \cite{perez2021rissanen} approximates hardness as the area under the curve of the test loss obtained by finetuning $\model{}$ on successively larger slices of the training data---a large area implies that a large number of training samples are required to achieve low test set loss, implying that the dataset is hard for $\model{}$.

\subsection{Our Approach} 

In the few-shot regime, one of the factors affecting dataset hardness is the degree to which input features for a given label in the test set differ from the training set (\figref{fig:spread_intuition}b). Based on this intuition, we propose a simple \emph{few-shot specific} hardness metric called {\spread{}} that computes the average euclidean distance of test input features to the closest training set input. 

For some input $z = (x, y) \in \mathcal{D}$, let $\modeln{z}$ denote the vector valued features of the input $x$ produced by the model. We define the distance between some example $z^{(k)}$ to the training set $\train$ as,
\begin{align}
    d(\model, z^{(k)}, \train) &\triangleq \min_{z} \|\modeln{z^{(k)}} - \modeln{z} \|  \\
    \textrm{subject to: } &z = (x,y) \in \train \nonumber\\
    &y = y^{(k)}. \nonumber   
\end{align}

Then, $\smetric{\model{}, \mathcal{D}}{\text{\spread{}}}$ can be computed as
\begin{align}
    \smetric{\model, \mathcal{D}}{\text{\spread{}}} &\triangleq \frac{1}{\lvert \mathcal{D}_{ts} \rvert}\sum_{z^{(k)} \in \test} d(\model, z^{(k)}, \train)
\end{align}

\subsection{Experiments} \label{sec:experiments}

\begin{table}[t]
\small
    \centering
    \begin{tabular}{lrr}
    \toprule
        Metric & Correlation & Compute Time (s)  \\
         \midrule
         $\msa{}$ & 0.06 & 80 \\ 
         $\mrda{}$ & 0.36 & 1,020 \\ 
        \midrule
         $\mspread{}$ & \textbf{0.47} & 10 \\ 
         \bottomrule
    \end{tabular}
    \caption{Compared to baseline hardness metrics, $\mspread{}$ is able to better account for intrinsic few-shot hardness and is computationally lightweight. All experiments are run on a single 1080ti GPU for profiling.}
    \label{tab:metrics}
\end{table}

We experiment with RDA, SA and {\spread{}} as our few-shot hardness metrics. To obtain input features for computing {\spread{}}, we use SimCSE \citep{gao2021simcse} features of the input with a RoBERTa-large base. Given a collection of datasets and methods, we measure the correlation between metric outputs and $\ifh$ values for each dataset, and report the average across all datasets. We also report time taken for computing the metric on a single 1080ti GPU for each dataset, and report the average compute time across all datasets. 

\paragraph{Results.} We report results on {\fsnli{}} to ensure uniformity of task formats and performance metrics. From Table~\ref{tab:metrics}, we observe that $\msa{}$ is poorly correlated with $\ifh$. Next, we note that while $\mrda{}$ produces better hardness judgements, these come at the cost of increased computation time. Finally, $\mspread{}$ is able to best account for intrinsic few-shot hardness while being \textasciitilde{}100x computationally lightweight compared to $\mrda{}$.

\subsection{Measuring $\ifh$ across pre-trained models} \label{sec:analysis_backbone}

While we define $\ifh$ of a dataset with respect to a fixed pre-trained model, certain datasets or tasks might be few-shot hard for a wider range of pre-trained models. To investigate this further, we experiment with Electra-large~\cite{clark2020electric} as the base pre-trained model. We obtain hardness measurements on {\fsnli{}} and compute the correlation between methods with different base pre-trained models (RoBERTa-large vs Electra-large). From average correlation across datasets in Table~\ref{tab:backbone_results}, we find that hardness measurements from the same method with different pre-trained models are well-correlated. We conclude that there may even be ``pre-trained model independent'' factors behind intrinsic few-shot hardness, and we leave further analysis of this to future work.

\begin{table}[t]
\small
    \centering
    \begin{tabular}{lr}
    \toprule
    
        Method &  Correlation  \\
         \midrule
         LMBFF &  0.72  \\
         AdaPET & 0.58 \\
         Null Prompt & 0.67\\
         Prompt-Bitfit & 0.79  \\
         \bottomrule
    \end{tabular}
    \caption{Correlation between hardness measurements from using the same adaptation method with different pre-trained models. We observe a high correlation among  hardness measurements from using different base pre-trained models.}
    \label{tab:backbone_results}
\end{table}

\section{Decreasing Few-Shot hardness via dataset decomposition}

\begin{table}[t]
\small
    \centering
    \begin{tabular}{lrrr}
    \toprule
    
        Method &  Ours & Control  \\
         \midrule
         LMBFF &  17.1 & \textbf{19.7}  \\
         AdaPET & \textbf{32.3} & 8.5  \\
         Null Prompt & \textbf{3.3} & -7.7 \\
         Prompt-Bitfit & \textbf{2.9} & -9.0 \\
         \midrule
         Few-shot Avg. & \textbf{13.9} & 2.9 \\
         \bottomrule
    \end{tabular}
    \caption{Comparing a {\spread}-inspired training set sampling strategy to random sampling. We observe an average boost of 13.9 accuracy points compared to the control where the \textasciitilde{}3 point boost is due to ensembling.}
    \label{tab:data_spread}
\end{table}

We conclude experiments with a simple application of {\spread{}} by proposing a training set sampling strategy that minimizes {\spread{}} values for improved few-shot performance. Our strategy is based on performing a ``clustering based decomposition'' of a dataset, similar to \citet{murty-etal-2021-dreca}. In particular, we run $k$-means clustering on input features from the pre-trained model to create $P$ clusters. Then, we train $P$ \emph{distinct} models on few-shot training sets sampled from each of the clusters. At test time, examples are classified into one of the $P$ clusters, and the corresponding model is used to make predictions. To account for any effects due to ensembling, we compare with an approach where we randomly sample a model to make predictions, instead of using the model corresponding to the cluster.

\paragraph{Results.} We experiment with the sampling strategy described above on a named entity classification based dataset from {\fsnli{}}. From Table~\ref{tab:data_spread}, we note an improvement of \textasciitilde{}14 accuracy points, while the control increases by only a \textasciitilde{}3 accuracy points.

\section{Related Work and Discussion}
Most notions of what makes datasets challenging are either model agnostic---example length \citep{spitkovsky-etal-2010-baby}, order sensitivity \citep{Nie2019AnalyzingCO} etc, or indirect \citep{DBLP:journals/corr/abs-2008-11600}. On the other hand, intrinsic hardness as defined in this work, is both model-centric and directly measures the ability of a \emph{specific} model family to make good generalizations from a training set. While measuring hardness of datasets has seen some recent traction \citep{10.1162/tacl_a_00403, perez2021rissanen}, to the best of our knowledge, we are the first to study hardness in the context of few-shot adaptation. We show that few-shot hardness of datasets (as measured by test set performance normalized against a random baseline) among an extensive set of recently proposed methods is highly correlated, suggesting that few-shot hardness may be a property intrinsic to datasets. We then propose a simple hardness metric called {\spread{}}, based on the intuition that a test set with input features close to the few-shot training set is easy, since it allows a model to exploit feature-space invariances. Compared to other metrics, {\spread{}} provides hardness judgements that are much better correlated with intrinsic few-shot hardness while being 8--100x faster to compute, compared to prior hardness metrics.

Metrics for predicting fewshot hardness have several applications. For the NLP practitioner, {\spread{}} could be used as a simple plug-and-play estimator of whether few-shot adaptation of a pre-trained model might be successful for a given use case. For the dataset developer, {\spread{}} could be used to adversarially curate harder few-shot benchmarks. Finally, a model developer can use {\spread{}} to inform better training set sampling strategies to improve test set performance in the few-shot training regime, as well as for model selection by selecting models with lower {\spread{}}.

\section{Acknowledgements}

SM was funded by a gift from Apple Inc. We are grateful to John Hewitt, Eric Mitchell, Roy Schwartz and the anonymous reviewers for helpful comments.

\section{Reproduciblity}
Our code is available at: \url{https://github.com/colinzhaoust/intrinsic_fewshot_hardness}.

\section{Limitations} 
\paragraph{Instance-level Analysis.} We focus on discovering and measuring intrinsic few-shot hardness at the dataset level and do not study \emph{instance-level} hardness quantitatively or qualitatively. Understanding hardness at the instance level can further help understand the recent successes behind few-shot learning in NLP.

\paragraph{Base model selection.} We compare few-shot performance between a variety of methods with two base models that have very different pre-training objectives, yet have correlated few-shot behaviors across tasks and adaptation methods. Of course, there exist a much wider range of pre-trained models with very diverse pre-training data and objectives. While beyond the scope of this work, we believe that studying factors such as the relationship between few-shot performance and pre-training data / objectives is worth further investigation.

\bibliography{anthology,custom}
\bibliographystyle{acl_natbib}

\clearpage

\appendix

\section{Appendix}

\begin{table}[t]
\small
    \centering
    \begin{tabular}{lrrr}
    \toprule
    
        Dataset &  \#Test & Majority \\
        \midrule
        event & 4,342 & 0.50 & \\
        ner & 37,638 & 0.50 & \\
        gender & 464 & 0.50 & \\
        puns & 1,756 & 0.50 & \\
        lexico\_syntactic & 15,236 & 0.51 & \\
        relation\_extraction & 761 & 0.60 & \\
        sentiment & 600 & 0.50 & \\
        semantic\_role & 1,821 & 0.59 & \\
        paraphrase & 2,109 & 0.55 & \\
        anaphora & 146 & 0.51 & \\
        negation & 1,000 & 0.67 & \\
        boolean & 1,000 & 0.73 & \\
        quantifier & 1,000 & 0.66 & \\
        counting & 1,000 & 0.66 & \\
        conditional & 1,000 & 0.66 & \\
        comparative & 1,000 & 0.65 & \\
        monotonicity & 2,000 & 0.67 & \\
        monotonicity\_simple & 1,000 & 0.67 & \\
        monotonicity\_hard & 1,000 & 0.68 & \\
        rte & 277 & 0.53 & \\
        mnli & 10,000 & 0.63 & \\
        
        \midrule
        ner\_merged & 36,789 & 0.51 & \\
        ner\_person & 9,032 & 0.55 & \\
        ner\_entity & 5,086 & 0.58 & \\
        ner\_location & 8,958 & 0.64 & \\
        ner\_event & 96 & 0.5 & \\
        ner\_organization & 7,851 & 0.5 & \\
        ner\_time & 5,766 & 0.65 & \\
         \bottomrule
    \end{tabular}
    \caption{Statistics of NLI datasets evaluated in this work, all the datasets have two label classes (entailed and not-entailed). The number of examples in support and validation set are jointly 64/128 per label class, except from \textit{ner\_merged}, where we use the support/test examples from all other \textit{ner\_x} tasks. \textit{ner} denotes the named entity classification (NEC) task and \textit{ner\_x} denotes the NEC task with \textit{x} as the label.}
    \label{tab:nli_stats}
\end{table}
\subsection{Dataset Details}
\label{sec:dataset_details}

We consider 11 tasks from the GLUE \cite{wang-etal-2018-glue} and SuperGLUE \cite{NEURIPS2019_4496bf24} benchmarks, namely SST-2~\cite{socher-etal-2013-recursive}, CoLA~\cite{warstadt2018neural}, MNLI~\cite{N18-1101}, QNLI ~\cite{rajpurkar-etal-2016-squad}, RTE~\cite{dagan_dolan_magnini_roth_2010}, MRPC~\cite{dolan-brockett-2005-automatically}, QQP~\cite{ijcai2017-579}, BoolQ~\cite{clark-etal-2019-boolq}, CB~\cite{cb_2019}, COPA~\cite{roemmele2011choice}, and WiC~\cite{pilehvar-camacho-collados-2019-wic}. We exclude the datasets that contain passages longer than the prompts.

We also take NLI datasets from \citet{white-etal-2017-inference,poliak-etal-2018-collecting, DBLP:journals/corr/abs-1909-07521}, giving us a collection of 28 datasets. If the original dataset contains three labels (entailment, contradiction, neutral), we merge contradiction and neutral to form a binary classification with entailed and not-entailed as the examples. The statistics of the datasets are shown in Table~\ref{tab:nli_stats}.

\subsection{Implementation Details}
\label{sec:hyperparam_details}

\noindent\textbf{Hyper-parameter Details:}
For grid search, we choose a learning rate from the set {5e-6, 1e-5, 5e-5} and train for 30 epochs. For each method, we train for 1000 steps, evaluating every 100 steps. All adaptation methods use RoBERTa-large as the base model. All other hyperparameters are as used in the original works. We follow the original work to generate and select prompts for LMBFF. For AdaPET, we use the same prompts as we used in LMBFF.

\begin{figure}

\centering
\includegraphics[width=\linewidth]{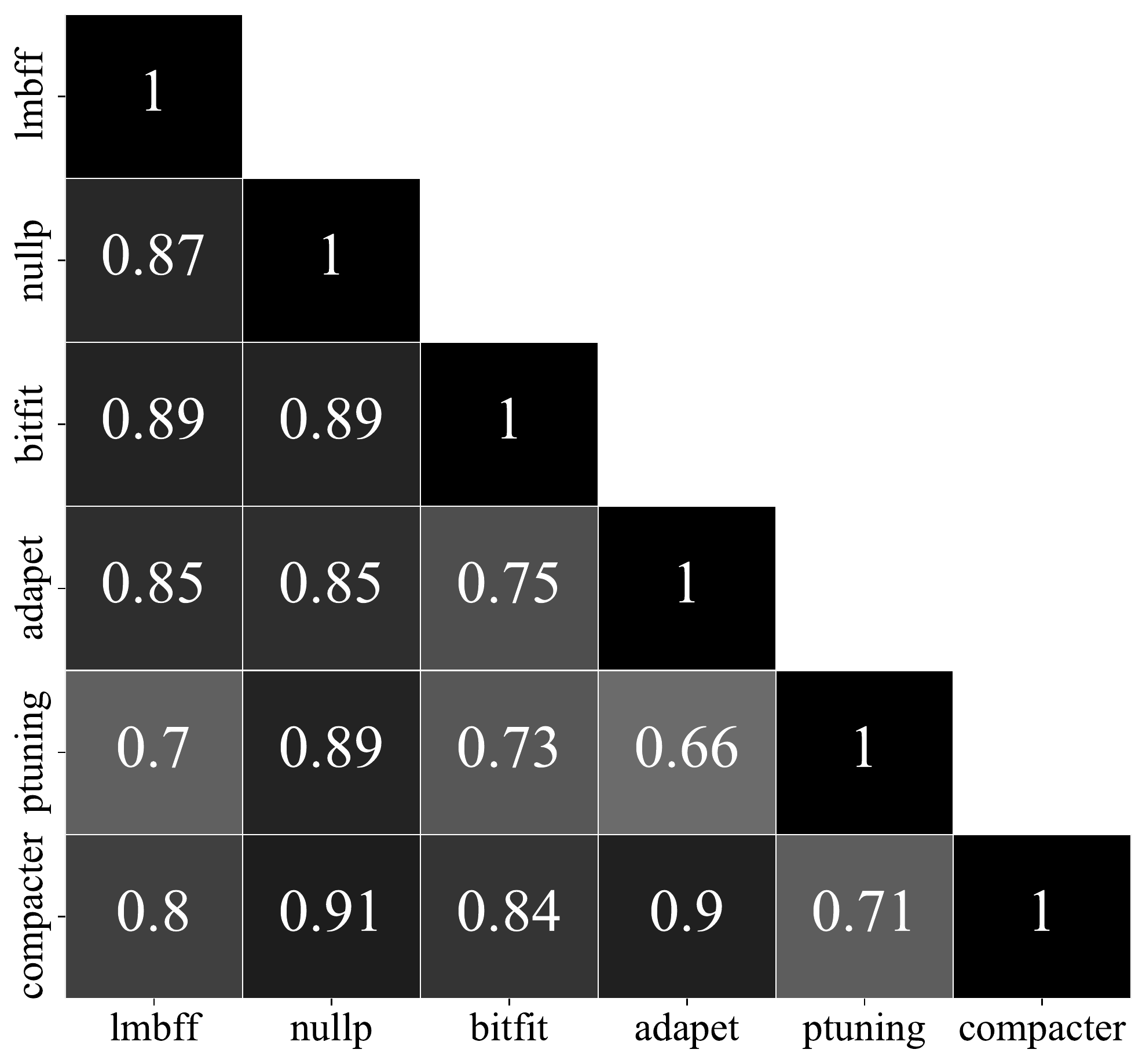}

\caption{Average correlation between hardness measurements from various few-shot adaptation methods on {\fsglue{}}. We observe general good correlation among various few-shot methods.}
\label{fig:intrinsic_heatmap_glue_method}
\vspace{-0.2in}
\end{figure}

\begin{figure}

\centering
\includegraphics[width=\linewidth]{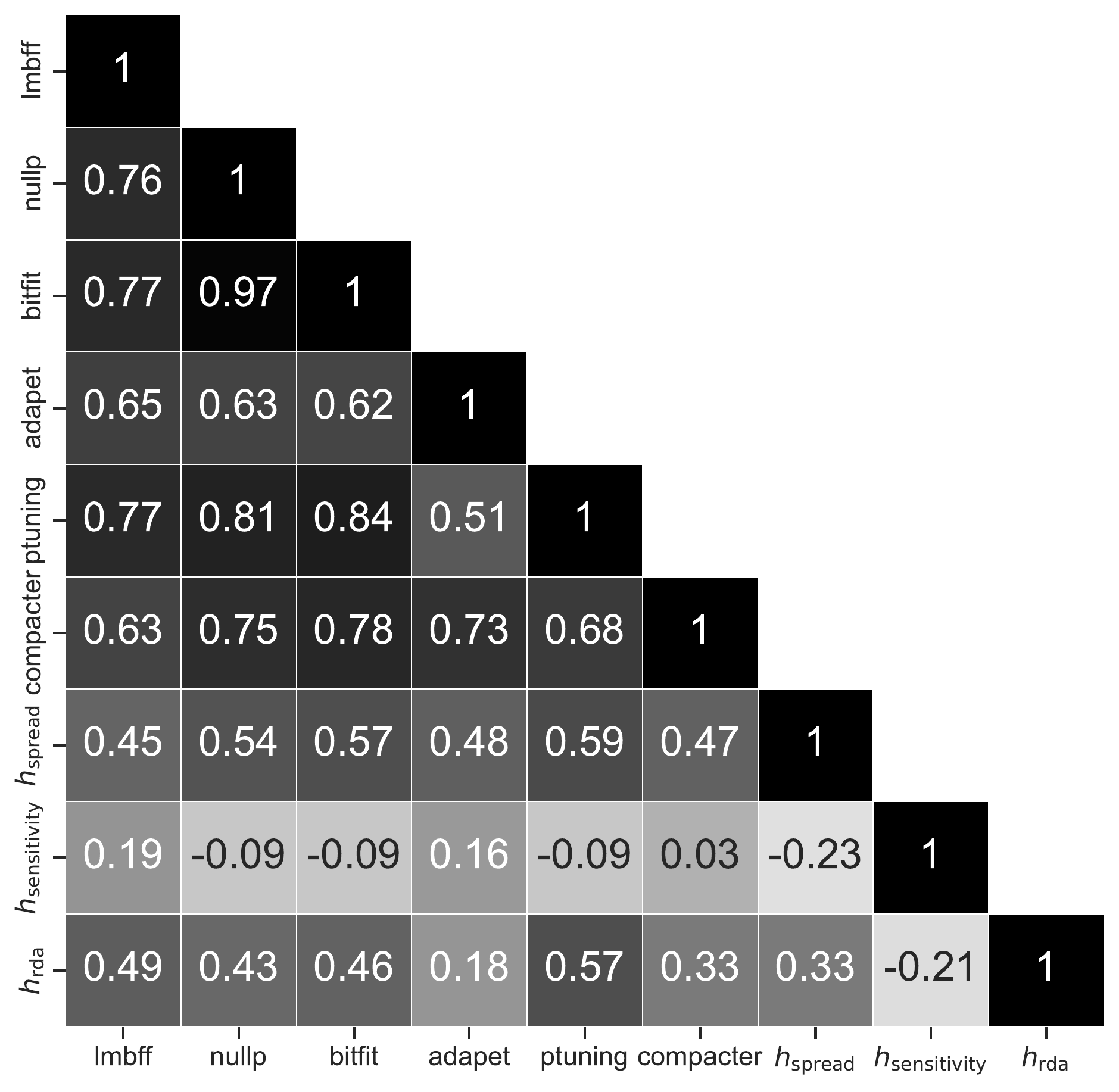}

\caption{Average correlation between hardness measurements from various few-shot adaptation methods on {\fsnli{}}. We observe general good correlation among various few-shot methods.}
\label{fig:intrinsic_heatmap_nli_method}
\end{figure}

\subsection{Intrinsic few-shot hardness}
In \figref{fig:intrinsic_heatmap}, we show averaged spearman correlations of methods clustered according to their type. In \figref{fig:intrinsic_heatmap_glue_method} and \figref{fig:intrinsic_heatmap_nli_method}, we further show detailed correlations between every pair of methods. We  generally observe high correlations across methods for both {\fsglue{}} and {\fsnli{}} tasks. In \figref{fig:intrinsic_heatmap_nli_method}, we further add $\mspread{}$, $\mrda{}$, and $\msa{}$ as references.

\subsection{Decreasing Few-Shot hardness via dataset decomposition (Addendum)}
We present the strategy of clustering based decomposition on named entity classification task from {\fsnli{}} and demonstrates its effectiveness in Table~\ref{tab:data_spread}. Since there naturally exist heuristics based cluster (e..g., questions about location, organization, and person) in NEC, we here show the relevance between model-generated clusters and these heuristics based cluster by computing the Jaccard index. From Table~\ref{tab:cluster_jaccard}, we can observe that model-based clusters are not decomposing the dataset in a similar way as the heuristics.

\begin{table}[t]
\small
    \centering
    \begin{tabular}{@{}lrrrrrr@{}}
    \toprule
    
       Clusters & 1 & 2 & 3 & 4 & 5 & 6\\
        \midrule
        person & 0.21 & 0.05 & 0.06 & 0.11 & 0.05 & 0.19 \\
        entity & 0.07 & 0.14 & 0.12 & 0.09 & 0.09 & 0.02\\
        location & 0.04 & 0.17 & 0.12 & 0.11 & 0.05 & 0.26 \\
        event & 0.03 & 0.03 & 0.01 & 0.03 & 0.02 & 0.02\\
        organization & 0.06 & 0.12 & 0.11 & 0.07 & 0.05 & 0.26\\
        time & 0.20 & 0.07 & 0.05 & 0.15 & 0.05 & 0.02\\
         \bottomrule
    \end{tabular}
    \caption{Jaccard index between each pair of clusters, where 1 to 6 denote the 6 clusters generated by k-means clustering over the representation space and person, entity, and etc denote the heuristics based clusters.}
    \label{tab:cluster_jaccard}
\end{table}

\end{document}